\pdfoutput=1

\documentclass[11pt]{article}

\usepackage[preprint]{acl}

\usepackage{times}
\usepackage{latexsym}

\usepackage[T1]{fontenc}

\usepackage[utf8]{inputenc}

\usepackage{microtype}

\usepackage{inconsolata}

\usepackage{graphicx}
\usepackage[nolist]{acronym}
\usepackage{multirow} 
\usepackage{amssymb} 
\usepackage{cancel}

\newcommand{\y}[0]{\checkmark}
\newcommand{\n}[0]{\texttimes}
\newcommand{\da}[0]{$\blacklozenge$}

\begin{acronym}
    \acro{rq}[RQ]{research question}
    \acroplural{rq}[RQs]{research questions}
    \acro{nlp}[NLP]{natural language processing}
    \acro{plm}[PLM]{pre-trained language model}
    \acroplural{plm}[PLMs]{pre-trained language models}
    \acro{llm}[LLM]{large language model}
    \acroplural{llm}[LLMs]{large language models}
    \acro{cs}[CS]{Conversational search}
    \acro{css}[CSS]{conversational search system}
    \acroplural{css}[CSSs]{conversational search systems}
    \acro{serp}[SERP]{search engine result page}
    \acroplural{serp}[SERPs]{search engine result pages}
    \acro{ai}[AI]{artificial intelligence}
    \acro{qa}[QA]{question-answering}
    \acro{kbqa}[KBQA]{question-answering over knowledge bases}
    \acro{kg}[KG]{knowledge graph}
    \acroplural{kg}[KGs]{knowledge graphs}
\end{acronym}

\title{Engineering Conversational Search Systems: A Review of Applications, Architectures, and Functional Components}

\author{Phillip Schneider$^1$, Wessel Poelman$^2$, Michael Rovatsos$^3$, and Florian Matthes$^1$ \\
         $^1$Technical University of Munich, Department of Computer Science, Germany \\
         $^2$KU Leuven, Department of Computer Science, Belgium \\
         $^3$The University of Edinburgh, School of Informatics, United Kingdom \\
         \texttt{\{phillip.schneider, matthes\}@tum.de} \\
         \texttt{wessel.poelman@kuleuven.be} \\
         \texttt{michael.rovatsos@ed.ac.uk}}
\begin{document}
\maketitle
\begin{abstract}
Conversational search systems enable information retrieval via natural language interactions, with the goal of maximizing users' information gain over multiple dialogue turns. The increasing prevalence of conversational interfaces adopting this search paradigm challenges traditional information retrieval approaches, stressing the importance of better understanding the engineering process of developing these systems. We undertook a systematic literature review to investigate the links between theoretical studies and technical implementations of conversational search systems. Our review identifies real-world application scenarios, system architectures, and functional components. We consolidate our results by presenting a layered architecture framework and explaining the core functions of conversational search systems. Furthermore, we reflect on our findings in light of the rapid progress in large language models, discussing their capabilities, limitations, and directions for future research.
\end{abstract}

\section{Introduction}

\label{sec:intro}
Accessing information has always been one of the primary functions of computer systems. Early systems relied on command-line interfaces with a specific syntax for data retrieval. As search systems evolved, database query languages enabled more complex queries but required technical knowledge. Then, free-text search engines allowed users to enter keywords in natural language, with information typically displayed as a result page listing relevant items \cite{hochstotter2009what}. In recent years, the evolution of search systems has continued in the direction of human-like dialogues.

Conversational search has emerged as a novel search paradigm, marking a shift from traditional search engines to interactive dialogues with intelligent agents \cite{radlinski2017theoretical,zhang2018conversational}. 
Many people have grown accustomed to using conversational interfaces like chatbots and voice assistants \cite{klopfenstein2017rise}. The widespread usage of dialogue systems has changed how humans expect to interact with computers \cite{mctear2016conversational}. Although modern conversational agents have impressive skill sets, their information-seeking capabilities are relatively limited and often confined to answering simple questions. As a consequence, there is a growing research interest in developing conversational search interfaces that go beyond simple query-response interactions by supporting more complex mixed-initiative dialogues, which is further fueled by the surging popularity of \acp{llm} and their integration into many kinds of search applications. 

Even though the topic of conversational search is relatively new, its fundamental concepts can be traced back to early works from the \ac{nlp} and information retrieval fields. So far, this emerging topic has been approached from different angles. While some researchers focus on theories and conceptual aspects \cite{azzopardi2018conceptualizing}, others conduct dialogue analyses and build prototypes to ground abstract models in empirical studies \cite{vakulenko2021largescale}. Yet, despite the ample literature about required properties, many proposed systems are too complex to implement. This apparent gap highlights the need for a more holistic inspection that connects theoretical requirements with realizable functional components.

We conducted a systematic literature review investigating different aspects of \acp{css} to address this research gap. The three main contributions are as follows: \newline (1) We identify the conceptual system properties and suitable application scenarios of \acp{css}. \newline (2) We consolidate architectures from the literature into a layered architecture framework and elaborate on the core functional components of \acp{css}. \newline (3) We discuss the manifold implications for augmenting \acp{css} with \acp{llm}, highlighting their potential capabilities, limitations, and risks.

\section{Related Work}
\label{sec:related-work}
In the related research literature on systems for conversational information-seeking, three categories are usually distinguished: search, recommendation, and \ac{qa} \cite{zamani2023conversational}. As the name suggests, \acp{css} actively involve users in the search process. Through multi-turn dialogues, users enter queries, locate information, examine results, or refine their search goals. In contrast to search systems, recommender systems usually rely on data about user preferences and past interaction histories to help with decision-making by providing personalized recommendations. \ac{qa} systems have been an active area of research for many decades. Given a text corpus or knowledge base and a dialogue history, conversational \ac{qa} systems aim to find answers to natural language questions \cite{vakulenko2021question}. It is worth noting that the boundaries between conversational search, recommender, and \ac{qa} systems are blurred and overlap. Although surveys exist on the two latter system categories \cite{jannach2021survey,zaib2022conversational}, our literature review is dedicated to search-oriented conversational interfaces. 

Despite the growing body of research on conversational search, related work, such as surveys or systematic literature reviews, remains scarce. The few studies we found tend to have a narrow topic focus on certain application domains or challenges. For example, the survey from \citet{adatrao2022survey} gives an overview of conversational search applications in biomedicine. In a different study, \citet{keyvan2022how} address the challenge of dealing with ambiguous queries. Another literature study from \citet{gerritse2020bias} investigates problematic biases in personalized content that conversational search agents can exhibit. Yet another work by \citet{kiesel2021metainformation} is a comprehensive survey on meta-information in search-oriented conversations. 

To the best of our knowledge, we are the first to provide a system-centric review across the development process, ranging from conceptualizing core functions to implementing architectural components. 
Unlike the mentioned studies, we do not look into specific challenges or domains within conversational search but take on a broad engineering perspective. We summarize valuable insights regarding the design and development of \acp{css} for several application use cases. Additionally, we address the recent interest surrounding \acp{llm} and their potential implications for engineering \acp{css}.

\section{Method}
\label{sec:method}
We conducted our systematic review based on the guidelines from \citet{kitchenham2004evidencebased}. Our study aims to shed light on the complex engineering process behind \acp{css} from initial system requirements to technical implementations by focusing on three key aspects: (1) definitions and proposed application scenarios to conceptualize the functional requirements of \acp{css}, (2) architectural elements suggested in the literature to effectively support these required system properties, and (3) core functions of \acp{css} discussed in the academic literature along with their implementations.

To obtain relevant publications, we devised a search string for querying six academic databases, as presented in Table~\ref{tab:method} of Appendix~\ref{sec:appendix-a}. The publication period was restricted to the time window between 2012 and 2022, yielding 212 candidate papers that predated the emergence of primarily \ac{llm}-based dialogue systems like ChatGPT \cite{openai2022chat}. Two researchers screened the papers for relevance, selecting a final set of 51 papers. Additionally, they performed forward and backward snowballing to include recent papers from 2023 and 2024, mainly focusing on \acp{llm} for \acp{css}.

\section{Results}
\label{sec:results}

\subsection{Definitions and Application Scenarios}
The concept of conversational search is not uniformly defined in the literature. We found three main categories of definitions. System-oriented definitions describe conversational search referring to architectural components \cite{sa2020challenges,vakulenko2021largescale}. Dialogue-oriented definitions emphasize the specifics of the dialogue interaction \cite{radlinski2017theoretical,kiesel2021metainformation}. Task-oriented definitions state tasks the system must complete \cite{zhang2018conversational,trippas2020model}. Despite focusing on different aspects, the analyzed definitions point out similar qualities to distinguish \acp{css} from traditional search approaches. These qualities are often related to the theoretical framework of \citet{radlinski2017theoretical}, which provides a structure and set of characteristics for designing and evaluating \acp{css}. In summary, we identified four reoccurring system properties from the analyzed papers. Firstly, \textit{mixed-initiative interaction} lets both user and system collaboratively steer the dialogue. Secondly, \textit{mutual understanding} involves the system revealing its capabilities and helping users express their needs. Thirdly, \textit{context awareness and memory} refers to the system's ability to gather information from its surroundings and conversation history to adapt dynamically. Lastly, \textit{continuous refinement} denotes improving retrieval performance through direct feedback or learning from past interactions.

\paragraph{Search Modality.}
These system properties open up a wide range of use cases, but the suitability of conversational search depends on the search modality and search task. \acp{css} can support text-based, speech-based, or hybrid interaction modalities. \citet{aliannejadi2021analysing} analyze various modality types and discuss their impact on the user's information gain during conversations. The authors mention examples like voice interfaces as speech-only options for service hotlines, text-based systems that can be integrated into messaging platforms or web search engines, and multimodal systems, such as virtual assistants or smart speakers with screens to display visual information. Contrary to text-based interfaces, spoken \acp{css} work without screens and are highly accessible because they do not require any technical expertise. Yet, conveying search results solely through speech output can overwhelm users \cite{deldjoo2021multimodal}. Moreover, two studies conducted by \citet{xing2022agerelated} and \citet{sa2020challenges} indicate that different modalities influence the search behavior concerning the frequency of query reformulation or how long search results are examined. Although the majority of \acp{css} in the literature are predominantly uni-modal and text-based, \citet{liao2021mmconv} note a growing trend towards multimodal systems.

\begin{figure*}[ht]
\centering
  \includegraphics[width=0.99\textwidth]{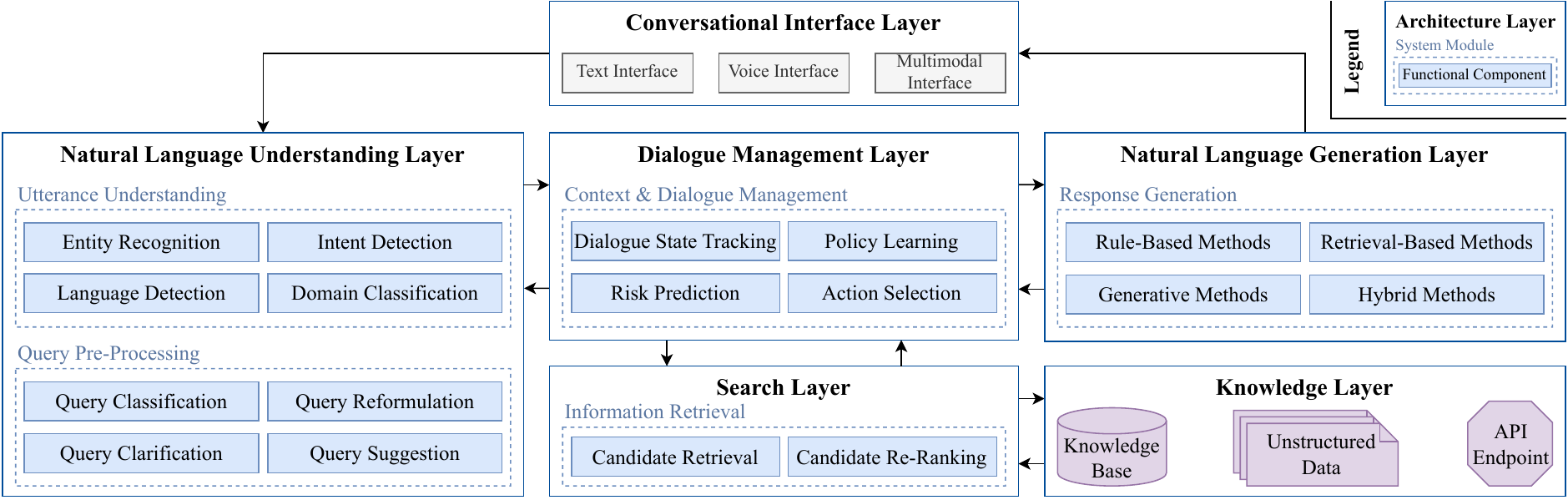}
  \caption{{Architectural framework of conversational search systems.}}
  \label{fig:architecture}
\end{figure*}

The modality and the nature of the search task determine the appropriateness of conversational interaction. A conventional data lookup with a graphical user interface may be more efficient in scenarios where the information need can be easily expressed. Concerning more ambiguous scenarios where the search goal is multi-faceted, and the data structure complex, a free-form conversation with iterative clarifications, reasoning steps, and feedback loops becomes applicable for conversational search \cite{radlinski2017theoretical}. In support of this, \citet{ren2021wizard} and \citet{schneider2023investigating} argue that dialogue-based search is particularly effective for exploratory search goals that involve progressively narrowing down information items from unfamiliar information spaces \cite{white2009exploratory}. Other tasks for which the usefulness of conversational search was highlighted are sequential \ac{qa}, learning about a new topic, asking for personal recommendations, or making plans \cite{anand2020conversational}.

\paragraph{Application Scenarios.} In our analysis of conversational search scenarios, we identified several real-world application domains that have been explored. While business and health were the most popular domains, we observed a significant diversification in the last years, including aerospace, gastronomy, law, news media, public services, or tourism \cite{liao2021mmconv}. For example, several researchers have studied product search in e-commerce scenarios for eliciting user preferences across multiple dialogue turns \cite{bi2019conversational,xiao2021endtoend}. A study from \citet{bickmore2016improving} proposed a \ac{css} to support people with low health and computer literacy to find information about clinical trials for which they may be eligible. In the domain of news media, \citet{schneider-etal-2023-data} demonstrate the integration of knowledge graphs with conversational interfaces to enhance exploratory search of newspaper articles. They present a knowledge-driven dialogue system and, through a large-scale user study with 54 participants, evaluate its effectiveness and derive design implications regarding functional improvements. \citet{liu2021conversational} compared conversational versus traditional search in a legal case retrieval scenario, showing that users achieve higher satisfaction and success in the conversational approach, especially when they lack sufficient domain knowledge. We find that the analyzed domain-specific systems often help overcome the absence of prior background knowledge, facilitating users in initiating the search process. Alternatively, these systems can provide assistance when the interface's modality is restricted and does not support conventional search methods.

\subsection{Architecture Framework}
Once the application scenario and desired system requirements are defined, the subsequent steps in the engineering process are to transform theoretical properties into technical implementations. This refers to functional components and their integration as part of the system architecture. We identified over 20 system architectures from the literature and consolidated reoccurring elements into the generalized \ac{css} architecture displayed in Figure~\ref{fig:architecture}. The proposed architecture adopts a layered architecture pattern, where each of the six layers performs a specific role within the \ac{css}. The layers contain modules and functional components specifically designed for information-seeking purposes. For example, the \textit{conversational interface layer} establishes the interaction channel between the system and the user. It receives user requests and presents search results depending on the modality. The three layers of \textit{natural language understanding}, \textit{dialogue management}, and \textit{natural language generation} deal with processing input utterances, handling conversation logic, and producing responses as output. In \acp{css}, the correct understanding and meaningful pre-processing of user queries are essential to maximize the information gain. The \textit{search layer}, in conjunction with the \textit{knowledge layer}, performs search operations within the information space, ensuring access to various data structures. Possible data sources are corpora with unstructured text documents, application programming interfaces (APIs), or structured knowledge bases like knowledge graphs \cite{schneider-etal-2022-decade}. Data items can be stored in various databases, such as relational, graph, or vector databases, each with distinct benefits and drawbacks based on the data characteristics and application needs.

\begin{table*}
\centering
\footnotesize
\begin{tabular}{p{2.6cm}p{3.2cm}p{4.6cm}p{2.2cm}c}
    \hline
    \textbf{Functions} & \textbf{Example Studies} & \textbf{Datasets} & \textbf{Models} & \textbf{Access} \\
    \hline
    \multirow{2}{*}[0.5em]{Query classification} & \citet{aliannejadi2020harnessing} & TREC CAsT & BERT &  \href{https://github.com/aliannejadi/castur}{\da} \\
     & \citet{voskarides2020query} & TREC CAsT, QuAC & BERT & \href{https://github.com/nickvosk/sigir2020-query-resolution}{\checkmark} \\
    \multirow{2}{*}[0.5em]{Query reformulation} & \citet{zhang2021chatty} & TREC CAsT & HWE, T5 & \href{https://github.com/castorini/chatty-goose}{\checkmark} \\
     & \citet{yu2020fewshot} & TREC CAsT & GPT-2 & \href{https://github.com/thunlp/ConversationQueryRewriter}{\checkmark} \\
    \multirow{2}{*}[0.5em]{Query clarification} & \citet{zamani2020generating} & Bing search logs & BiLSTM & \n \\
     & \citet{bi2021asking} & Qulac & BERT & \href{https://github.com/aliannejadi/qulac}{\da} \\
    \multirow{2}{*}[0.5em]{Query suggestion} & \citet{rosset2020leading} & Bing search logs & BERT, GPT-2 & \n \\
     & \citet{mustar2022study} & TREC Session, MARCO, AOL logs & BERT, BART, T5 & \da \\
    \multirow{2}{*}[0.5em]{Candidate retrieval} & \citet{xiong2020approximate} & TREC DL, NQ, TriviaQA & ANCE & \href{https://github.com/microsoft/ANCE}{\y} \\
     & \citet{lin2021contextualized} & TREC CAst, CANARD, MARCO & BERT & \da \\
    \multirow{2}{*}[0.5em]{Candidate re-ranking} & \citet{kumar2020making} & TREC CAsT & BERT & \da \\
     & \citet{mele2021adaptive} & TREC CAsT, ConvQ & BERT & \href{https://github.com/hpclab/adaptive-utterance-rewriting-conversational-search}{\y} \\
    \multirow{2}{\linewidth}{Knowledge-based response generation} & \citet{zhang2020summarizing} & WikiTableQuestions & T5, GPT-2 & \href{https://github.com/iai-group/sigir2020-tablesum}{\da} \\
     & \citet{ren2021conversations} & SaaC & PPG & \href{https://github.com/PengjieRen/CaSE_RG}{\y} \\
     \hline
    \end{tabular}
    \caption{Example studies, datasets, and implementations of the seven core functions in conversational search. Legend: $\checkmark = $ dataset(s) and system; $\blacklozenge = $ dataset only; $\times = $ not available.}
\label{tab:example-studies}
 \vspace{-0.2cm}   
\end{table*}

Modules group functional components and thus represent a specific functionality inside the layers. There is a separation of concerns among the modules, which deal only with logic pertinent to their respective layer. For instance, the query pre-processing module is a functionality from the language understanding layer, which enhances user queries through reformulation, clarification, suggestion, or other functions. The components perform specific tasks on the lowest abstraction level using \ac{nlp} techniques. Implementing a component usually requires training \ac{nlp} models that receive an input and classify, retrieve, or generate textual data, in some instances also structured data. Components can be implemented independently, requiring knowledge only of how they are connected to other components. While the displayed architecture encompasses all components encountered in the literature, implementations of a concrete \ac{css} usually employ only a subset of these components. For example, reacting to user feedback is an essential function often mentioned in theoretical frameworks, but only a few studies implement it as part of an actual system \cite{bi2019conversational,wang2021controlling}. Since most architectures focus only on specific functional components like query suggestions or generating clarifying questions, there is a discrepancy between theoretical frameworks and practical implementations. Section~\ref{sec:functions} provides a more detailed overview of the various conversational search-specific core functions from the architectural components.

In line with common architectural patterns for dialogue systems, our proposed architecture follows a layered structure, separating functionality into different modules. We found that most analyzed implementations from the literature connect modules in a pipeline-based approach \citep[\textit{inter alia}]{rojasbarahona2019spoken,mele2021adaptive,alessio2023decaf}. However, we observed a growing number of research works aiming to develop end-to-end approaches with transformer-based neural networks instead of classic \ac{nlp} pipelines \cite{xiao2021endtoend,ferreira2022opendomain}. While end-to-end learning enables training a single model to represent target modules without the usual intermediate steps found in pipeline designs, these systems still depend on multiple task-specific modules and do not achieve a genuine end-to-end design, where only one model would handle all functionalities. To date, even the most advanced \acp{llm} fail to integrate all functions without encountering issues, as we will discuss in more depth later on. 

An example of a pipeline-based architecture is the open-source framework called \textit{Macaw} from \citet{zamani2020macaw}. It consists of three modules implemented in a generic form with replaceable \ac{nlp} models. One module is responsible for query pre-processing with co-reference resolution and query reformulation or expansion, another for ranking documents with a retrieval model, and a third module for response generation. Two system proposals from \citet{zhang2021chatty} and \citet{mele2021adaptive} have similar architectural components but additionally adopt a neural passage re-ranker for re-ordering results of the first-stage retrieval using a BERT model \cite{nogueira2019passage}. Concerning end-to-end approaches, \citet{xiao2021endtoend} introduce a \ac{css} for online shopping, consisting of a sequence-to-sequence transformer for dialogue state tracking and a multi-head attention mechanism to match user queries to products. Comparable architectures from \citet{ren2021conversations} and \citet{ferreira2022opendomain} that aim to implement conversational search sub-tasks in an end-to-end manner also include transformers, such as BERT and T5, for passage re-ranking and response generation models.

Our presented architecture framework captures the fundamental aspects of \acp{css} in the research literature, and although there might be architectural adaptations to suit specific application scenarios with varying interface modalities and data structures, the body of six layers remains unchanged. The architecture offers flexibility in adding, removing, or replacing components within the modules.

\subsection{Conversational Search Functions}
\label{sec:functions}
This section elaborates on the seven core functions of \acp{css} mentioned in the architecture framework. Implementing these functions using \ac{nlp} techniques is the most concrete step in the engineering process. Therefore, we review example studies that implement commonly used machine learning models (see Table~\ref{tab:example-studies}) and list the most popular training and evaluation datasets in Table~\ref{tab:datasets} of Appendix~\ref{sec:appendix-a}. Despite being essential for conversational systems, some components like intent detection are not explicitly explained here as they are not specific to \acp{css}. While not all functions may be present in a given system or are combined, these main functions have been widely utilized and are treated as individual sub-tasks in the broader fields of conversational search and information retrieval. The order of paragraphs for each function roughly follows the processing steps needed to generate an output given an input turn in the conversation.

\paragraph{Query Classification.}

As part of the initial query pre-processing module, classifying the given query can benefit many subsequent system components. In conversational search scenarios, user requests may not be self-explanatory and ambiguous due to a lack of context. Researchers have approached this problem by classifying what type of question is being asked \cite{kia2020opendomain}, determining the search domain of interest \cite{frummet2019detecting,hamzei2020place}, or deciding whether a (past) query is relevant in the context of the ongoing dialogue \cite{aliannejadi2020harnessing,voskarides2020query}.
Other system components can adapt according to classified queries, such as querying domain-specific sources, discarding irrelevant utterances, or selecting relevant past utterances. The often-used TREC Conversational Assistant Track (CAsT) datasets contain many sessions where a user inquires about two subjects and later asks questions to compare the two. Classification can be used to select the previous relevant utterances.  

\paragraph{Query Reformulation.}

Since a \ac{css} is processing dialogue turns, it has to deal with many subtleties and challenges. Conversational search primarily deals with ambiguity and co-reference issues \cite{keyvan2022how}.
Reformulating, also called rewriting, a query to an unambiguous and explicit form is often needed for effective information retrieval and to incorporate contextual information of an ongoing conversation.
Numerous approaches incorporate transformer-based language models for this task \cite{ferreira2022opendomain}. Either as a classifier to determine what terms have to be incorporated into the rewritten query \cite{mele2021adaptive}, a sequence-to-sequence approach trained on \mbox{\emph{query -- rewrite target}} pairs \cite{zhang2021chatty} or in a weakly-supervised fashion using \acp{llm} \cite{yu2020fewshot}. The following is a simple example of rewriting:
\noindent
\begin{tabular}{ll}
    User: & Who is the director of Citizen Kane? \\
    System: & Orson Welles is the director. \\
    User: & Does he have children? \\
    Rewrite: & Does \cancel{he} \emph{Orson Welles} have children? \\
\end{tabular}

\paragraph{Query Clarification.}

When the system cannot resolve or interpret a query, it can take the initiative and ask the user for clarification.
\acp{css} that can show initiative, such as proactively asking questions, are referred to as \textit{mixed-initiative} systems.
Different approaches for clarifying questions have been investigated, including template filling, sequence editing models, sequence-to-sequence models, and combinations of these methods.
Template filling can be as straightforward as \emph{``Did you mean $X$?''} for a misspelling or co-reference issue.
Templates can cover many clarifying questions, but their specificity level is something to consider \cite{zamani2020generating}.
Sequence editing models are related to query rewriting; they choose a clarification question and rewrite it with information from the ongoing dialogue state \cite{zamani2023conversational}. 
Sequence-to-sequence approaches train models with \mbox{\emph{unclear query -- clarifying question}} pairs to predict fitting questions.

Asking a clarifying question is not always the best course of action.
Systems have to ensure a user's patience or tolerance is not running out by asking too many questions \cite{bi2021asking}.
Controlling this `risk' and the system's information need is a delicate balance.
Current approaches implement functions that try to approximate the information gain and tolerance of a user \cite{salle2021studying,wang2022simulating}.
If the system wants to ask a clarifying question, it uses this function to decide whether it should proceed. This can be done for numerous reasons. \citet{braslavski2017what} provide a taxonomy of six clarification categories. Their categorical taxonomy is created from analyzing \emph{community question-answering} websites but can be applied more generally.

\paragraph{Query Suggestion.}

\acp{css} can help users while they are still in their conversational turn by suggesting relevant queries or even (partial) answers while the interaction is ongoing \cite{aliannejadi2021analysing,keyvan2022how}.
Search engines are a good example of this, where auto-complete is heavily used.
Suggesting queries can possibly mitigate issues addressed by the previously mentioned system functions.
If the system incorporates dialogue state information in the suggestions, it can provide unambiguous versions of an unclear query.
Generating query suggestions is done in many ways, but all must deal with the query, dialogue state, and ranking-generated suggestions.
An often-used approach is training a model to determine what to copy or generate from the dialogue state and input query to maximize the chance of a user picking the suggestion \cite{dehghani2017learning,mustar2022study}.
The generated queries can be ranked by the same or a separate model \cite{rosset2020leading}.

\paragraph{Candidate Retrieval.}

Candidate retrieval fetches possibly relevant data items by producing a structured database query given the (pre-processed) user query or retrieving information from unstructured text collections. The latter approach falls into two general categories: sparse retrieval and dense retrieval \cite{gao2023neural}.
Sparse retrieval ranks documents with methods such as BM25 \cite{robertson2009probabilistica}.
These use sparse vectors encoding term occurrences in queries and documents, which can be used for retrieval directly, to perform pre-filtering of results \cite{vakulenko2021question,zhang2021chatty}, or to represent model features (e.g., for re-ranking) \cite{cho2021personalized}.
Although computationally efficient, the purely lexical approach of these methods limits them in dealing with synonyms, word order, and spelling mistakes.

Dense retrieval addresses these issues, which is often implemented as a \emph{dual encoder} architecture, where one neural model encodes a document into a dense vector and another the (processed) query \cite{lin2021contextualized}.
These models are trained by jointly training these two encoders on labeled \mbox{\emph{query -- relevant document}} pairs.
There are variations with additional encoding strategies, but the main idea stays the same \cite{ferreira2022opendomain}.

\paragraph{Candidate Re-Ranking.}                                                         

Once the system has a set of possibly relevant candidate results for the current turn or utterance, the next step is to rank this set in order of informativeness.
There are many approaches to re-ranking, with the most dominant one being some model that either classifies, scores, or re-orders a given input set \cite{ferreira2022opendomain}.
These models are either fine-tuned on explicitly labeled \mbox{\emph{query -- relevant item}} pairs \cite{zhang2021chatty,mele2021adaptive} or use some distance measure between (part of) the embedded query and (part of) the relevant document.
These are the main building blocks of most implementations, but they can be combined into more elaborate setups.
\citet{kumar2020making}, for instance, suggest \emph{multi-view re-ranking}, where the system creates different embeddings of the input query.
These \emph{views} include information from dialogue history, relevant terms from the retrieved items, and the rewritten query, which get fused into the final ranking.

\paragraph{Knowledge-Based Response Generation.}

The final step of a turn in the conversational system is to present the response to the user in the form of natural language.
As with information retrieval, natural language generation is a dedicated research field.
As such, many distinct approaches and methods within \acp{css} exist.
These are generally grouped according to three categories: the information type, generation method, and information source.

Information type refers to the response's structure based on the retrieved document(s) or information need.
These include \emph{short answer}, \emph{long document retrieval}, \emph{abstractive summarization} or \emph{structured entities} \cite{zamani2023conversational}.
For instance, a short factual question often does not require a large response (\emph{``In what year did $X$ happen?''}). In contrast, a query for an explanation might involve summarizing a relevant passage.

Different generation methods are used for these different answer types and can serve as a grouping of approaches.
Some general methods include; template filling \cite{zhang2018conversational}, sequence-to-sequence methods \cite{ferreira2022opendomain} and weakly supervised approaches \cite{baheti-etal-2020-fluent}.
More elaborate approaches have a model choosing from where to copy a token in generating the response: a vocabulary, the input query, or the retrieved passage \cite{ren2021conversations,ren2021wizard}.

Generation is also dependent on the information source being queried.
Conversational search is generally done over a corpus of free text but can also be done over a knowledge graph \cite{kacupaj2022contrastive,dutt2022perkgqa} or other (semi-)structured information \cite{zhang2020summarizing}.
The source influences the choice of generation technique; verbalizing a sub-graph from a knowledge graph is considerably different from summarizing a text passage.

There are also hybrid methods that fuse information sources and generation methods.
The most influential contribution in this area has been \emph{retrieval-augmented generation} \cite{lewis2020retrieval,shuster-etal-2021-retrieval-augmentation}.
These hybrid approaches try to balance the expressiveness and veracity of responses.

\section{Discussion and Future Directions}

\label{sec:discussion}
The results from our review give insights into the engineering behind \acp{css} from abstract properties to realizable functional components. Against this background, our findings unveil a disruptive trend of adopting larger language models to integrate end-to-end functional components. Researchers have emphasized the benefits of streamlined \ac{nlp}, reduced error propagation, and data-driven development. Hence, rather than reflecting on the numerous general challenges in the evaluation of \acp{css}, like \citet{penha2020challenges}, we direct our focus toward discussing how \acp{llm} can augment \acp{css} and the implications it has on their future evolution.

While most studies fine-tune language models (e.g., BERT or T5) on downstream tasks, there has been a recent surge of interest in using \acp{llm}. By scaling up models to billions of parameters and training them on corpora with trillions of tokens, \acp{llm} have demonstrated emergent capabilities and prowess in multi-task learning \cite{radford2019language}. A significant advantage of \acp{llm} is prompt-based (or in-context) learning. Through carefully defined prompts, \acp{llm} can perform multiple tasks without specific training or tuning \cite{pengfei2023}. Furthermore, there has been a growing interest in optimizing \acp{llm} for dialogue interactions by pre-training on conversations, instruction fine-tuning, and reinforcement learning from human feedback \cite{thoppilan2022lamda}. The strengths of \acp{llm}, such as their language understanding and ability to generate context-aware responses, make them highly complementary elements for \acp{css}.

\paragraph{Opportunities for Conversational Search.}
A rapidly growing body of new studies concentrates on advancing conversational search functions with \acp{llm}. For instance, addressing the challenge of better understanding user queries, \citet{anand2023query} introduce a query formulation framework to replace multi-component pipelines with a single \ac{llm}. This model initially generates several machine intents for a user query, followed by options to accept, edit, or expand these intents until they align with the user's query intent. With a qualitative feasibility study, the authors show that the \ac{llm}-generated rewrites can improve the downstream retrieval performance. In related work, \citet{mao2023large} investigate different prompting and aggregation methods for performing few-shot conversational query reformulation with \acp{llm}. They demonstrate that their approach outperforms state-of-the-art baselines by testing a GPT-3 model on CAsT'19 and '20 datasets. Another study from \citet{chen2023graph} introduces a retrieval-based query rewriting approach, where an \ac{llm} leverages external knowledge from graphs with historical user-entity interactions and collaborative filtering. \citet{ye-etal-2023-enhancing} also demonstrate the potential of \acp{llm} for query rewriting, showing that rewrites can significantly enhance retrieval performance in conversational search. Furthermore, \acp{llm} can augment \acp{css} through semantic parsing and convert a natural language question into a structured database query. For example, \citet{schneider2024evaluating} evaluate how well different-sized \acp{llm} perform in generating knowledge graph queries for conversational \ac{qa} based on dialogues by comparing various prompting and fine-tuning techniques. Aside from query rewriting and semantic parsing, \acp{llm} can also be effective for classifying query intents \cite{srinivasan-etal-2022-quill} or generating clarification questions \cite{kuhn2023clam}. 

In addition to the natural language understanding layer, \acp{llm} can augment the layers of dialogue management, search, and natural language generation. For example, \citet{friedman2023leveraging} developed a system for conversational video search and recommendation powered by several \acp{llm} based on the LaMDA model \cite{thoppilan2022lamda}. While one \ac{llm} is used as a dialogue management module, a second \ac{llm} acts as a re-ranker module. This \ac{llm} also generates explanations for its decisions. The authors discuss how a third \ac{llm} can be instructed to act as a user simulator for generating synthetic data for training and evaluation. Also focusing on synthetic data generation, a paper from \citet{huang-etal-2023-converser} introduces a framework called \textit{CONVERSER} that uses \acp{llm} to generate conversational queries given a passage in a retrieval corpus for training dense retrievers. This can significantly benefit conversational search by reducing the need for extensive and expensive data collection while maintaining high retrieval accuracy. Concerning knowledge-based text generation, \acp{llm} have also proven to be effective for verbalizing semantic triples retrieved from graph-structured data, with performance improvements achievable through few-shot prompting, post-processing, and fine-tuning techniques \cite{schneider-etal-2024-comparative}. Another noteworthy approach from \citet{sekulic2024towards} employed \acp{llm} in conversational search for answer rewriting, proposing two strategies by either providing inline definitions of important entities or offering users the opportunity to learn more about entities. Human-based evaluations indicated a preference for the answers with inline definitions.

\paragraph{Challenges and Risks.}
Even though \acp{llm} show great potential for conversational search, they have known shortcomings that must be considered. First, the sheer size of these models requires significant computational resources. Multiple graphical processing units are often necessary for enabling fast inference, a critical factor for conversational search applications that require responses in near real-time. The research community has been actively exploring solutions such as model distillation, model quantization, or low-rank adaptation to address these issues. Distillation involves compressing \acp{llm} into smaller and more efficient versions \cite{shridhar-etal-2023-distilling}. Model quantization is a technique where the floating point precision of model parameters is decreased, leading to smaller memory requirements and faster computations without significant performance loss \cite{xiao2023smooth}. Low-rank adaptation fine-tunes only a subset of the model's parameters rather than updating the entire parameter space \cite{hu2022lora}.

Other major issues related to \acp{llm} are hallucinating or omitting important information and a lack of transparency regarding the source from which the output was generated \cite{dou-etal-2022-gpt,ji2023survey,xu2024hallucination}. To mitigate these risks, scholars have looked into approaches to ground the generated outputs in trustworthy data sources and mechanisms to curate generated output. For example, \citet{peng2023check} introduce a framework for augmenting \acp{llm} by first incorporating retrieved evidence from external knowledge as input context and then using \ac{llm}-generated feedback as instructions to revise responses. Through validation with two information-seeking tasks, the authors show that their approach reduces hallucinations while preserving fluency and usefulness. Another knowledge-enhancement method from \citet{wang2023empower} fine-tunes a smaller \ac{llm} (Llama-7B) to learn domain-specific knowledge. This model is consulted to generate expert opinions that are used to enrich the prompt context of a bigger, general \ac{llm} (GPT-4) to improve its domain-specific \ac{qa} capabilities. For a comprehensive survey of over 30 hallucination mitigation techniques, readers are referred to \citet{tonmoy2024comprehensive}. Regardless, it must be noted that \acp{llm} are nondeterministic by nature, making it challenging to ensure consistent and persistent knowledge during searches due to the inherent randomness in their text generation methods \cite{krishna-etal-2022-rankgen,pmlr-v202-mitchell23a}.

Finally, there are efforts to develop software tools that address the reliability and safety of generated \ac{llm} output by adding programmable guardrails as well as logical control patterns. Popular tools that aid the development of \ac{llm}-based \acp{css} include \textit{NeMo} \cite{nvidia2023nemo}, \textit{Guidance} \cite{microsoft2023guidance}, and \textit{LangChain} \cite{chase2022langchain}. Other tools like \textit{DeepEval} \cite{deepeval2024} can evaluate model bias, which is crucial since \acp{llm} in conversational search can increase selective exposure and opinion polarization by fostering confirmatory querying behaviors \cite{sharma2024generative}. In summary, ongoing research shows the potential of \acp{llm} to advance the engineering of dialogue-based search systems with various approaches to mitigate their reliability issues. However, it is unlikely that \acp{llm} will replace \acp{css} as a single end-to-end monolith in the foreseeable future. Instead, they are more likely to augment the modular structure of the proposed architecture framework.

\section{Conclusion}
\label{sec:conclusion}
We conducted a comprehensive review of engineering \acp{css}, establishing connections between theoretical application scenarios and technical implementations. Based on our analysis of existing architectures, we introduced a layered architecture framework and explained its functional core components. While it is essential to acknowledge that the field of conversational search is rapidly evolving, and complete coverage is unattainable, our framework provides a generalized architecture based on previously validated systems. The framework does not claim to be exhaustive but rather serves as a foundational starting point for designing and developing \acp{css}. Lastly, we discussed recent work on the capabilities and challenges of augmenting \acp{css} with \acp{llm}. We outline where they fit into our proposed framework, which core functions they have been used for, and highlight promising directions for future research.

\section*{Acknowledgements}
This work has been supported by the German Federal Ministry of Education and Research (BMBF) Software Campus grant 01IS17049.

\bibliography{custom}

\newpage
\onecolumn
\appendix
\section{Appendix}
\label{sec:appendix-a}

The Appendix provides supplementary material for our study, including a list of the six queried academic databases along with the applied search string (Table~\ref{tab:method}), as well as an overview of commonly used datasets for \acp{css} (Table~\ref{tab:datasets}).

\begin{table*}[h]
\footnotesize
    \centering
    \begin{tabular}{lcl}
    \hline
    \multicolumn{3}{l}{\textbf{Search String}} \\ \hline
    \multicolumn{3}{l}{
    \begin{tabular}[c]{@{}l@{}}
    \textit{``conversational search''} OR\\
    \textit{``information-seeking dialogue''} OR\\ 
    \textit{``conversational information retrieval''} OR\\
    \textit{``conversational information-seeking''} OR\\
    \textit{``information-seeking conversation"}\\
    \end{tabular}
    } \\ \hline
    \textbf{Database} & \textbf{Number of Papers} & \textbf{Database Link} \\ \hline
    ACL Anthology & 48 & \url{https://aclanthology.org} \\ 
    ACM Digital Library & 101 & \url{https://dl.acm.org} \\ 
    IEEE Xplore & 5 & \url{https://ieeexplore.ieee.org/Xplore} \\ 
    ScienceDirect & 3 & \url{https://www.sciencedirect.com} \\ 
    Scopus & 46 & \url{https://www.scopus.com} \\ 
    Web of Science & 9 & \url{https://www.webofscience.com/wos/} \\ \hline
    \end{tabular}
    \caption{Search string and number of retrieved candidate papers per database.}
    \label{tab:method}
\end{table*}

\begin{table*}[h]
\centering
\footnotesize
\begin{tabular}{p{6cm}p{2.4cm}p{5cm}c}
\hline
\textbf{Dataset} & \textbf{Size} & \textbf{Source} & \textbf{Lang.} \\
\hline
Amazon Reviews \cite{ni-etal-2019-justifying} &  9M products & Amazon product catalog & en \\
CANARD \cite{Elgohary:Peskov:Boyd-Graber-2019} & 40K questions & QuAC dataset & en \\
CodeSearchNet \cite{husain2019codesearchnet} & 2M code queries & GitHub repositories & en \\
ConvQ \cite{christmann2019convq} & 11K QA dialogues & Wikipedia & en \\
DuConv \cite{wu2019proactive} & 30K dialogues & MTime.com & zh \\
MRQA \cite{fisch2019mrqa} & 550K QA pairs & 18 existing QA datasets & en\\
MS MARCO \cite{nguyen2016ms} & 1M QA pairs & Bing search engine & en \\
MSDialog \cite{qu2018analyzing} & 2K QA dialogues & Microsoft Community forum & en \\
Natural Questions \cite{kwiatkowski2019natural} & 320K QA pairs & Google search engine & en \\
QuAC \cite{choi2018quac} & 14K QA dialogues & Wikipedia & en \\
Qulac \cite{aliannejadi2019asking} & 10K QA pairs & TREC Web Track & en \\
SaaC \cite{ren2021conversations} & 748 QA pairs & TREC CAR, MS MARCO, WaPo news & en \\ 
TREC CAR \cite{dietz2017trec} & 30M passages & Wikipedia & en \\
TREC CAsT \cite{dalton2020trec} & 38M passages & TREC CAR, MS MARCO & en \\
TriviaQA \cite{joshi2017triviaqa} & 650K QA pairs & Wikipedia, quiz and trivia websites & en \\
WikiTableQuestions \cite{pasupat-liang-2015-compositional} & 22K QA pairs & Wikipedia & en \\
\hline
\end{tabular}
\caption{Commonly used datasets in the literature on conversational search systems.}
\label{tab:datasets}
\end{table*}

\end{document}